\documentclass[runningheads]{llncs}
\usepackage{graphicx}
\usepackage{marvosym}
\usepackage{epstopdf}
\usepackage{float}
\usepackage{caption}
\usepackage{booktabs}
\usepackage{multirow}
\usepackage{xurl}
\usepackage{amsmath}
\usepackage{amssymb}
\usepackage{array}
\usepackage{bm}
\usepackage{makecell}
\usepackage{verbatim}
\usepackage{hyperref}
\usepackage{ifsym}

\begin{document}
\title{Promoting Open-domain Dialogue Generation through Learning Pattern Information between Contexts and Responses}
\titlerunning{Promoting Open-domain Dialogue Generation}
%

\author{Mengjuan Liu$^{(\textrm{\Letter})}$ \and Chenyang Liu    \and
Yunfan Yang \and Jiang Liu \and   Mohan Jing  }

\authorrunning{Mengjuan Liu et al.}
\institute{Network and Data Security Key Laboratory of Sichuan Province, University of Electronic Science and Technology of China, Chengdu, China \
\email{ mjliu@uestc.edu.cn, 202121090104@std.uestc.edu.cn} 
}
\maketitle              
\begin{abstract}
Recently, utilizing deep neural networks to build the open-domain dialogue models has become a hot topic. However, the responses generated by these models suffer from many problems such as responses not being contextualized and tend to generate generic responses that lack information content, damaging the user’s experience seriously. Therefore, many studies try introducing more information into the dialogue models to make the generated responses more vivid and informative. Unlike them, this paper improves the quality of generated responses by learning the implicit pattern information between contexts and responses in the training samples. In this paper, we first build an open-domain dialogue model based on the pre-trained language model (i.e., GPT-2). And then, an improved scheduled sampling method is proposed for pre-trained models, by which the responses can be used to guide the response generation in the training phase while avoiding the exposure bias problem. More importantly, we design a response-aware mechanism for mining the implicit pattern information between contexts and responses so that the generated replies are more diverse and approximate to human replies. Finally, we evaluate the proposed model (RAD) on the Persona-Chat and DailyDialog datasets; and the experimental results show that our model outperforms the baselines on most automatic and manual metrics.

\keywords{open-domain dialogue, pre-trained language model, exposure bias, response-aware mechanism}
\end{abstract}
\section{Introduction}
Natural language dialogue has long been one of the most challenging AI problems, and it is usually divided into two categories: task-oriented dialogue and open-domain dialogue \cite{gao2019neural}. Compared with the task-oriented dialogue system, it is more challenging for an open-domain dialogue system to ensure the quality of responses since their dialogue contents are not restricted, making it not easy to design response templates in advance. As a result, current open-domain dialogue systems are mainly built based on deep neural networks which support generating responses for various dialogue contexts \cite{huang2020challenges}. However, there is a general problem with existing models that tend to generate more generic responses that lack informativeness and are not contextually relevant \cite{li2015diversity}, such as “I don’t know.” and “I am not sure.” Such replies are so tedious that they easily discourage users from continuing the conversations, although they are grammatically correct. \par

To address this problem, some studies have attempted to make better use of the information already available, such as the ground-truth responses labeled in the training samples. In these studies, teacher-forcing  \cite{williams1989learning} is a representative technology to improve model performance using ground-truth responses. When using teacher-forcing, the input to the dialogue model at each decoding step is not the word generated at the previous decoding step but the corresponding word in the labeled response. Teacher-forcing can effectively prevent inappropriate words generated at the previous decoding steps from misleading the word generation at the subsequent decoding steps, thus improving the performance and learning efficiency of the dialogue model.\par


On the other hand, the existing dialogue models mainly focus on the semantic information of the context and ignore the learning of the hidden pattern information between the context and the response, which makes the responses generated by the models not fit well with the real responses. In this paper, we try to make the generated responses more interesting and vivid, similar to human responses, by introducing the implicit pattern information of the context-response pairs into the pre-trained dialogue model. This new response-aware dialogue model is named RAD. \par

Unlike common dialogue models, RAD improves the quality of generated responses by learning the ground-truth response information. When training, it first uses an improved scheduled sampling method to reconstruct the response vector. To the best of our knowledge, it is the first study to improve the quality of generated responses by mining the implicit pattern information between contexts and responses. The contributions of this paper can be summarized as follows:
\begin{itemize}
\item {We present a basic scheme for performing dialogue generation using the pre-trained model, based on which we propose a scheduled sampling method oriented to pre-trained models. }
\item {We design a response-aware mechanism that includes a response-aware network and a response-aware prediction network. In the training phase, we use the response-aware network to learn a representation vector containing the implicit pattern information of the context-response and feed this vector into the pre-trained model. In the generation phase, we replace the original response-aware vector with the response-aware vector predicted by the response-aware prediction network. This design enriches the pre-trained model’s available information while avoiding model performance degradation in the generation phase due to exposure bias.}
\item {We have evaluated the proposed model (RAD) on the Persona-Chat \cite{zhang2018personalizing} and DailyDialog \cite{li2017dailydialog} datasets. The experimental results show that RAD performs better than the baseline models on most automatic and human evaluation metrics. Also, we upload the datasets, codes, and results to GitHub \footnote{https://github.com/RussellLiu0/RAD} to help researchers in related fields reproduce our experiments quickly.}

\end{itemize}

\section{Background}
\subsection{Open-domain Dialogue Generation}

Open-domain chatbots that do not restrict conversation contents have been a hot topic in NLP research. However, as mentioned in Section 1, whether the seq2seq model or the pre-trained model is used to generate responses, the quality of the generated responses is still not good enough compared with the real responses in the training sample, and the utterances generated by the model can hardly reach the richness and interestingness of human language. Some studies have tried to improve the quality of responses by introducing various external information, such as interlocutor personality \cite{liu2020you}, common-sense knowledge \cite{zhang2019grounded}, emotional information \cite{zhou2018emotional} and other additional supplementary information to improve the quality of the response. These practices have indeed made some progress, but they also have many limitations, such as obtaining additional information. Therefore, we want to make better use of existing information, such as real responses in the training sample, to improve the performance of the model by mining the implicit pattern information between the context and the responses. \par 
\subsection{Scheduled Sampling}
A widely used technique for resolving exposure bias is scheduled sampling, which means that at each decoding step, the model probabilistically selects the generated word to replace the corresponding word in the ground-truth response as the decoder input during training  \cite{bengio2015scheduled}, thus reducing the reliance of the model on ground-truth responses. At the beginning of training, the model selects more words from the response as inputs, and as the training epochs increase, the model selects more generated words as inputs. Many studies have verified that scheduled sampling can effectively address exposure bias. Furthermore, the authors in \cite{zhang2019bridging} improved the original scheduled sampling method from the perspective of sampling granularity. They proposed using both sentence and word sampling granularity to obtain candidate texts for model training. In \cite{xu2021adaptive}, Xu et al. proposed an adaptive switching mechanism that can learn to transition between ground-truth learning and generated learning depending on the word-level matching score, such as the cosine similarity. \par

Unfortunately, these scheduled sampling methods are all for seq2seq models and cannot be applied straightforwardly to pre-trained models. In the seq2seq model, the decoder’s input at each decoding step is determined step-by-step when training, however, in the pre-trained model, all words in the response part are required to feed into the model together, making it impossible to perform partial replacements with the generated words because no words have been generated before the response vector is fed into the model. Unlike existing solutions, we propose an improved scheduled sampling method by which teaching-forcing can be used in a pre-trained dialogue model to alleviate the exposure bias problem.\par
\section{Response-aware Dialogue Model}
\begin{figure*}[tbp]
\centering
\includegraphics[width=\textwidth]{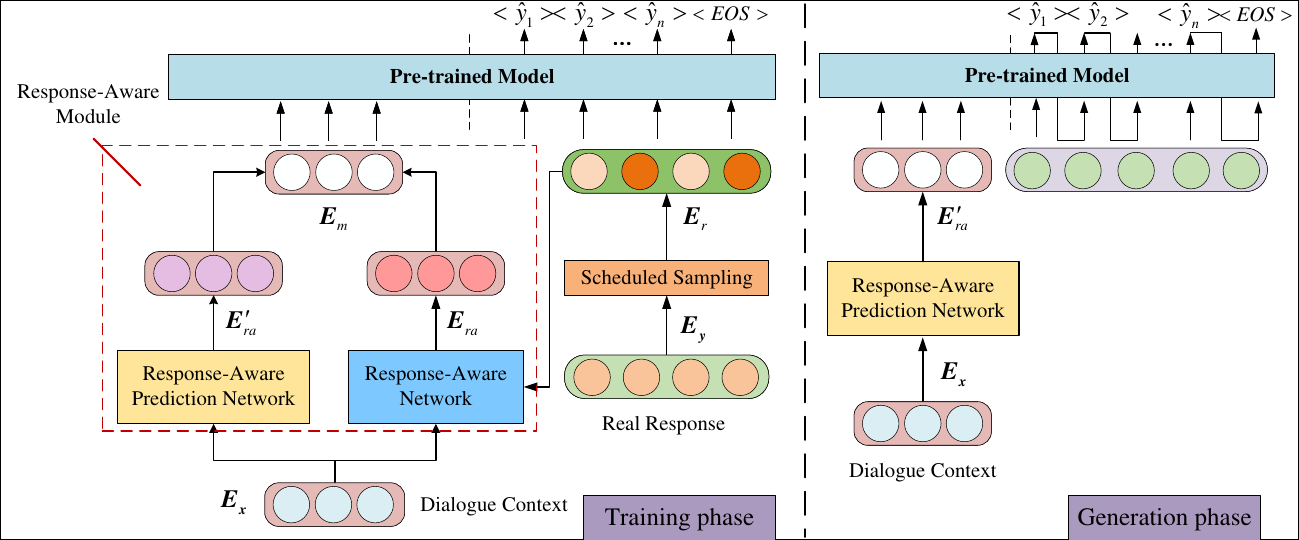}
\captionsetup{font={footnotesize}}
\caption{The structure of the RAD model (the left part is the structure for training and the right part is the structure for generating)}
\label{fig:fig1}
\end{figure*}

In this paper, we focus on the single-turn open-domain dialogue. The task of the dialogue model is to generate a syntax-correct, contextually relevant response according to the given context. Unlike the existing pre-trained dialogue models, the response-aware dialogue model (RAD) proposed in this paper adopts two differentiated structures in the training and generation phases. As shown in Figure \ref{fig:fig1}, the left part is used for training, and the right part is used for generation. \par

\subsection{Pre-trained Language Model}
This subsection describes how to use GPT-2, which only contains the Transformer decoder, to generate dialog responses. For a training sample, given the context $\bm{x}\!=\!\{ x_1,x_2,\cdots,x_{m} \}$ and the ground-truth response $\bm{y}\!=\!\{ y_1,y_2,\cdots,y_{n}\}$ , we first map them into two sequences of word embeddings, recorded as $\bm{E}_x=\{ \bm{e}_1^x,\bm{e}_2^x,\cdots,\bm{e}_{m}^x\}$ and $\bm{E}_y=\{ \bm{e}_1^y,\bm{e}_2^y,\cdots,\bm{e}_{n}^y\}$. Here, $\bm{e}_i^x$($\bm{e}_{i}^{x} \in \mathbb{R}^{1 \times L}$) denotes the word embedding of the $i$-th word in the context, $\bm{e}_j^y$($\bm{e}_{j}^{y} \in \mathbb{R}^{1 \times L}$) denotes the word embedding of the $j$-th word in the response, $L$ is the dimension of the word embedding, and $m$ and $n$ are the lengths of the context and response, respectively. Then, the concatenation of the two sequences is input into GPT-2 for computing the output vector corresponding to each response position, as shown in Equation (1), where $\bm{h}_i$ is the output vector of the $i$-th position. \par
\begin{equation}
\boldsymbol{H}=\operatorname{Pretrained}\left(\bm{E}_{x}, \bm{E}_{y}\right), \boldsymbol{H}\!=\!\{ \bm{h}_1,\bm{h}_2,\cdots,\bm{h}_{m},\bm{h}_{m+1},\cdots,\bm{h}_{m+n} \}
\end{equation}

Next, we transform each response position’s output vector into a probability distribution of vocabulary dimensions by a softmax function, as shown in Equation (2). Here, $\bm{P}_t$ is the generation probability distribution of the word corresponding to the $t$-th response position, $V$ is the vocabulary size, and $\bm{W}$ and $\bm{b}$ are two trainable matrices for the linear transformation.\par
\begin{equation}
\boldsymbol{P}_t=\operatorname{softmax}\left(\bm{W} \bm{h}_{m+t} + \bm{b}\right),
\boldsymbol{P}_t\!=\!\{ p_1^t,p_2^t,\cdots,p_{V}^t\},t\!=\!\left(1,2,\cdots,n\right) 
\end{equation}

Finally, we calculate the model’s loss function according to Equation (3), where $\bm{P}_t\left(y_t\right)$  denotes that, for the $t$-th response position, probability of generating the corresponding ground-truth response word $y_t$ according to $\boldsymbol{P}_t$. Like other pre-trained models, we fine-tune the parameters of GPT-2 by minimizing its loss. We should note that the response is produced word-by-word in the generation phase, so each response word in the training phase should be generated without reference to words in subsequent positions. Therefore, we use the attention mask mechanism to mask the attention scores of all words after the current response position in the training phase.\par
\begin{equation}
\operatorname{Loss}_{M}=-\sum_{t=1}^{n} \log \boldsymbol{P}_{t}\left(y_{t}\right) 
\end{equation}

%

\subsection{Scheduled Sampling for Pre-trained Model}

In this subsection, we improve the scheduled sampling method to make it available for the pre-trained model. Its primary task is reconstructing the response vector where some words in the real response are replaced by generated words. Specifically, the method consists of two stages. In the first stage, we input the context and response embeddings into the pre-trained model and get the generation probability distribution of the word corresponding to each response position, working as described in Section 3.1. Then, we average the embeddings of $K$ words with the highest generation probabilities at each response position $t$ to obtain its candidate embedding, as shown in Equation (4). Here, $\bm{e}_t$ is the candidate embedding vector of the $t$-th word in the response, and $\bm{e}_k^t$ is the word embedding with the $k$-th highest generation probability at the $t$-th response position. The standard scheduled sampling directly selects the word with the highest generation probability as the candidate word; however, the proposed scheduled sampling chooses the averaged word embedding as the candidate embedding vector. This approach smooths out the negative impacts of incorrectly generated words (caused by the greedy algorithm) on the results.\par

\begin{equation}
\boldsymbol{e}_{t}=\frac{1}{K} \sum_{k=1}^{K} \boldsymbol{e}^{t}_k,t\!=\!1,2,\cdots,n
\end{equation}

In the second stage, we determine whether to replace the word embedding at each response position with the candidate embedding according to the probability $p$. Referring to the literature \cite{zhang2019bridging}, the probability $p$ increases monotonically with the increase of training epochs and is calculated as shown in Equation (5). Here, $\mu$ is the hyper-parameter, and $l$ denotes the number of training epochs. Based on Equation (5), the probability increases more smoothly. In practice, we first generate a random number between 0 and 1 at each response position. If the random number is less than the probability $p$, we perform the replacement; otherwise, we still use the word embedding in the ground-truth response. Finally, we obtain the reconstructed response embedding sequence, $\boldsymbol{E}_{r}=\left(\boldsymbol{e}_{1}^{r}, \boldsymbol{e}_{2}^{r}, \cdots, \boldsymbol{e}_{n}^{r}\right)\left(\boldsymbol{E}_{r} \in \mathbb{R}^{n \times L}\right)$. It will be fed into the response-aware network to calculate the response-aware vector first and then fed into the pre-trained model for fine-tuning together with the response-aware vector. \par
\begin{equation}
\emph{p}=\frac{1}{1+(\mu / e^{l / \mu})}
\end{equation}
\subsection{Response-aware Mechanism}
In order to make better use of the ground-truth responses in the training samples, we design a response-aware mechanism for mining the implicit pattern information between contexts and responses. First, we design a response-aware network based on a multi-head attention mechanism \cite{vaswani2017attention} to compute the response-aware vector, as shown in Equation (6). Here, $\boldsymbol{E}_{x}$, $\boldsymbol{E}_{r}$ and $\boldsymbol{E}_{ra}$ are the context vector, the reconstructed response vector, and the response-aware vector. $\boldsymbol{E}_{ra}$ is a matrix that contains $m$ rows and $L$ columns with the same form as $\boldsymbol{E}_{x}$. Each row corresponds to a word embedding, and $L$ is the dimension of the word embedding.\par
\begin{equation}
\boldsymbol{E}_{ra}=\operatorname{MultiHead}\left(\boldsymbol{E}_{r}, \boldsymbol{E}_{x},\boldsymbol{E}_{x} \right)
\end{equation}

However, the responses are not available in the generation phase, so it is impossible to compute the response-aware vector. To this end, we design a response-aware prediction network to replace the response-aware network to estimate the response-aware vector, as shown in Figure \ref{fig:fig1}. The response-aware prediction network is implemented by a feedforward neural network that only takes the dialogue context as input and outputs the predicted response-aware vector, defined in Equation (7).\par
\begin{equation}
\boldsymbol{E}_{ra}^{\prime}=\operatorname{FeedForward}\left(\boldsymbol{E}_{x}\right)
\end{equation}

To make the estimated response-aware vector  $\boldsymbol{E}_{ra}^{\prime} \left(\boldsymbol{E}_{ra}^{\prime} \in \mathbb{R}^{m \times L}\right)$ approximate the response-aware vector $\boldsymbol{E}_{ra} \left(\boldsymbol{E}_{ra} \in \mathbb{R}^{m \times L}\right)$, we merge the output vectors of the response-aware network and the response-aware prediction network in the training phase, as shown in Equation (8), where $\lambda$ is a decreasing factor to balance the weights of the two components. We input the merged response-aware vector $\boldsymbol{E}_{m} \left(\boldsymbol{E}_{m} \in \mathbb{R}^{m \times L}\right)$ instead of the context’s embedding vector to the pre-trained model together with the reconstructed response vector $\boldsymbol{E}_{r}$.\par
\begin{equation}
\begin{gathered}
\boldsymbol{E}_{m}=\lambda \boldsymbol{E}_{ra}+(1-\lambda) \boldsymbol{E}_{ra}^{\prime}
\end{gathered}
\end{equation} 

Here, we illustrate the differences between the inputs of the pre-trained model in the training and generation phases by defining two equations. In Equation (9), the inputs of the pre-trained model are the merged response-aware vector $\boldsymbol{E}_{m}$ which contains the implicit pattern information of context-response, and the reconstructed response vector $\boldsymbol{E}_{r}$ obtained by the scheduled sampling module. In (10), the input of the pre-trained model is the predicted response-aware vector $\boldsymbol{E}_{ra}^{\prime}$ which is output by the response-aware prediction network only depending on the dialogue context.\par

\noindent 
In the training phase:
\begin{equation}
 \boldsymbol{H}=\operatorname{Pretrained}\left(\bm{E}_{m}, \bm{E}_{r}\right) 
\end{equation}
In the generation phase:\par
\begin{equation}
 \boldsymbol{H}=\operatorname{Pretrained}\left(\bm{E}_{ra}^{\prime}\right) 
\end{equation}  

In order to make the predicted response-aware vector approximate the response-aware vector, we additionally define a response-aware prediction loss, as shown in (11), which is used to measure the deviation between the predicted response-aware vector and the response-aware vector.\par
\begin{equation}
\operatorname{Loss}_{R A}=\frac{1}{m L} \sum_{i=1}^{m} \sum_{j=1}^{L}\left(e_{i j}^{\prime}-e_{i j}\right)^{2} ,\left(e_{i j}^{\prime} \in \boldsymbol{E}_{r a}^{\prime}, e_{i j} \in \boldsymbol{E}_{r a}\right)
\end{equation} 

Finally, the total loss function of RAD is a weighted sum of the pre-trained model’s loss and the response-aware prediction loss, defined in Equation(12), where $\gamma$ is a hyper-parameter that indicates the weights of the two losses.\par
\begin{equation}
\operatorname{Loss}_{\text {total}}=\gamma \operatorname{Loss}_{M}+(1-\gamma) \operatorname{Loss}_{RA}
\end{equation}

\section{Experiment Settings}

\paragraph{\textbf{Datasets.}}
We evaluate the proposed model (RAD) on two public datasets: Persona-Chat \cite{zhang2018personalizing} and DailyDialog \cite{li2017dailydialog}. Persona-Chat is a multi-turn dialogue dataset collected from a crowdsourcing platform, where volunteers conduct conversations following multiple simple personality descriptions. Each conversation has 5$\sim$8 turns with 4$\sim$6 persona descriptions. Another dataset, DailyDialog, is collected from an English learning website, in which the utterances come from the conversations between learners. DailyDialog is also a multi-turn dialogue dataset like Persona-Chat; the difference is that DailyDialog does not have explicit textual persona descriptions. As for the statistics of the dataset, Persona-Chat has 131431/7793 utterances pairs in the training/testing dataset, and DailyDialog has 76052/6740 utterances pairs in the training/testing dataset. \par

\paragraph{\textbf{Baselines.}}

As far as we know, our work is the first study to exploit labeled responses in the training samples to improve the quality of generated responses. Therefore we only compare our model with two basic models based on seq2seq and GPT-2. All the baseline models are described as follows:
\begin{itemize}
\item {\textbf{Seq2Seq} \cite{sutskever2014sequence}: a basic seq2seq dialogue model with an attention mechanism  \cite{vaswani2017attention}, where the encoder extracts contextual information and the decoder generates replies. Both encoder and decoder are built on single-layer LSTM networks.}
\item {\textbf{Seq2Seq+SS+RA}: a seq2seq dialogue model with the addition of scheduled sampling (SS) and response-aware (RA) mechanisms.}
\item {\textbf{GPT-2} \cite{radford2019language}: a dialogue model based on GPT-2, where we concatenate the context vector with the response vector and input them into the pre-trained model for generating the response. }
\item {\textbf{RAD (ours)}: a dialogue model based on GPT-2 with our scheduled sampling (SS) and response-awareness (RA) mechanisms. It is our model proposed in this paper. }

\end{itemize}

\paragraph{\textbf{Implement Details.}}
We build our model with pytorch. The pre-trained dialogue models are built based on GPT-2 \cite{radford2019language}, which contains 117M parameters, using 12 stacked Transformer layers. Each layer of the Transformer uses a multi-head attention mechanism containing 12 heads, and the dimension of the hidden layer is 768. The vocabulary contains 52370 words with BPE word encoding \cite{sennrich2015neural}, and the word embedding matrix comes from GPT-2. In the improved scheduled sampling method, the hyper-parameters $K$ and $\mu$ in equations (4) and (5) are set to 5 and 4. The response-aware network adopts a multi-head attention mechanism with eight heads, and the response-aware prediction network has one hidden layer with 768 neurons. The balance factor $\lambda$ in Equation (8) decreases quickly from 1 to 0.2 in the first training epoch and stays constant. The weight $\gamma$ of the pre-trained model’s loss in Equation (12) is set to 0.5. \par

We adopt the Adam optimizer to train all models. The learning rate of the seq2seq dialogue models is set to 1e-4, and the corresponding batch size is 32. The learning rate of the pre-trained dialogue models is set to 1e-5, and the batch size used in training is 16 since the pre-trained model has more parameters than the seq2seq model. This setting difference mainly takes into account the GPU performance. Furthermore, the optimizer does not have other special settings. \par

\section{Experimental Results}
\subsection{Automatic Evaluation}
In this subsection, we automatically evaluate the proposed and baseline models using metrics of F1, BLEU-1/2, and DISTINCT-1/2. BLEU and F1 are usually used to measure the similarity between generated and labeled responses. A model gets a higher BLEU/F1 score, indicating that it produces better responses than other models. DISTINCT \cite{li2015diversity} is an evaluation metric that measures the diversity of the generated results. A higher score on this item indicates better diversity in the responses generated by the model.\par
\begin{table}[htb]\Huge
\centering
\captionsetup{font={normal}}
\caption{Automatic evaluation results (best results are marked in \textbf{bold}).}
\resizebox{.95\textwidth}{!}{
\renewcommand{\arraystretch}{1.5}
\begin{tabular}{lccccccc}
\hline

\multirow{2}{*}{Model} & \multicolumn{3}{c}{Persona-Chat} & & \multicolumn{3}{c}{DailyDialog} \\
\cline{2-4} \cline{6-8}
 & {F1} & {BLEU-1/2} & {DISTINCT-1/2} &  & {F1} & {BLEU-1/2} & {DISTINCT-1/2} \\
\hline
{Seq2Seq} & {16.551} & {0.156/0.052} & {0.007/0.017} & & {16.589} & {0.108/0.041} & {0.027/0.117} \\
{Seq2Seq+SS+RA} & {16.554} & {0.163/0.056} & {0.005/0.010} & & {16.689} & {0.121/0.047} & {0.031/0.144}  \\
{GPT-2} & {18.790} & {0.177/0.083} & {\textbf{0.022}/0.092} & & {19.095} & {0.146/0.067} & {0.042/0.195} \\
{RAD(ours)} & {\textbf{19.385}} & {\textbf{0.190/0.089}} & {\textbf{0.022/0.093}} & & {\textbf{24.656}} & {\textbf{0.244/0.140}} & {\textbf{0.043/0.249}} \\
\hline
\end{tabular}}%
\label{tab:tab2}%
\end{table}%
Table \ref{tab:tab2} shows the results of the automatic evaluation. First, two pre-trained dialogue models (GPT-2 and RAD) have substantial improvements in all metrics over two seq2seq dialogue models, which proves that the pre-trained language models can indeed increase the quality of the responses generated by the models. Second, RAD achieves the best performance among all models. On the Persona-Chat dataset, compared with the basic GPT-2, RAD improves 3.17\% on the F1 score and 7.34\%/7.23\% on the BLEU-1/2, demonstrating that learning the context-response pattern information can help to increase the model performance. Finally, the basic seq2seq model obtains the lowest F1 and BLEU scores. However, adding the scheduled sampling method and the response-aware mechanism to the seq2seq model, the two metrics improve, especially the BLEU scores, which increased by 4.49\% for BLEU-1 and 7.69\% for BLEU-2. It shows that the two proposed improvements also effectively increase the response quality generated by the traditional dialogue model. \par

On the DailyDialog dataset, RAD has also achieved good results in all metrics. Compared to GPT-2, RAD has improved by 29.12\% on the F1 score and 67.1\%/108.9\% on the BLEU-1/2 scores, respectively. To our surprise, RAD also shows good improvements in the DISTINCT-2 scores. The results mean that the proposed response-aware mechanism helps enhance the similarity between generated and labeled responses and positively improves the diversity of generated responses. Overall, the two improvements proposed in this paper achieve consistent enhancements on both datasets with seq2seq and pre-trained models. \par

\subsection{Human Evaluation}
Generally, automatic evaluation is deficient in measuring the responses’ fluency and semantic consistency, so we tried to conduct human evaluation to assess the responses generated by the model in terms of fluency, contextual consistency, and similarity. First, we randomly sampled 100 testing samples from the Persona-Chat and DailyDialog datasets. Then, the responses generated by the models and reference responses are provided to five reviewers. Finally, reviewers scores each response according to the follow rules without knowing the response is generated by which model. A score of 2 means that the model performs well and generates responses that meet the requirements, and a score of 0 means that it performs poorly. The results are summarized in Table \ref{tab:tab4}. In this table, fluency is defined as whether the response is fluent and natural, coherence is defined as whether the response is relevant and coherent to the context and similarity is defined as whether the response is close to the real response. \par

\begin{table}[htb]\Huge

\centering
\captionsetup{font={normal}}
\caption{Human evaluation results (best results are marked in \textbf{bold}).}
\resizebox{.95\textwidth}{!}{
\renewcommand{\arraystretch}{1.5}
\begin{tabular}{lccccccc}
\hline
\multirow{2}{*}{\huge Model} & \multicolumn{3}{c}{Persona-Chat} & & \multicolumn{3}{c}{DailyDialog} \\
\cline{2-4} \cline{6-8}
 & {Fluency} & {Coherence} & {Similarity} & & {Fluency} & {Coherence} & {Similarity} \\
\hline
{Seq2Seq} & {1.823} & {0.825} & {0.455} & & {\textbf{1.946}} & {0.771} & {0.386} \\
{Seq2Seq+SS+RA} & {1.828} & {0.423} & {0.431}&  & {1.914} & {0.852} & {0.454}  \\
{GPT-2} & {\textbf{1.865}} & {1.172} & {0.815} & & {1.933} & {1.293} & {0.680} \\
{RAD(ours)} & {1.858} & {\textbf{1.475}} & {\textbf{0.940}} & & {1.916} & {\textbf{1.362}} & {\textbf{0.896}} \\

\hline
\end{tabular}}%
\label{tab:tab4}%
\end{table}%

First, we can see that on the Context Coherence metric, the scores of the two pre-trained dialogue models have been significantly improved over the two seq2seq-based dialogue models, indicating that the pre-trained models outperform the earlier seq2seq models in terms of context understanding. Further, RAD obtains higher Context Coherence scores than GPT-2, which indicates that the proposed response-aware mechanism does help the model to generate responses more consistent with the dialogue context. Finally, on the similarity metric, RAD improves the similarity with human responses by mining the pattern information compared with the original GPT-2. It should be noted that the results’ Fleiss’ kappa score \cite{fleiss1973equivalence} is 0.494 on the Persona-Chat dataset and 0.413 on the DailyDialog dataset, indicating reviewers agree consistently on the scores.\par

\subsection{Ablation Study}
In addition, we perform ablation experiments to validate the effectiveness of the proposed mechanisms. First, we observe the impacts of two improvements on the performance of the pre-trained dialogue models. Compared to the basic GPT-2, the scores obtained by GPT-2+SS are decreased in all metrics, indicates that replacing the real response vector with the reconstructed response vector input to the pre-trained model during training have negative impacts on the model performance. Similar to seq2seq models, the performance of GPT-2+RA that adds the response-aware mechanism to the basic GPT-2 has been significantly improved, which proves the effectiveness of RA. Further, the F1 and BLEU scores of GPT-2+RA+SS have been increased on the Persona-Chat dataset but decreased on the DailyDialog dataset. Through ablation experiments, we verify the effectiveness of our improvements on the model performance. Clearly, the performance improvements by the RA mechanism are significant, while the performance improvements of SS are not as well as expected. Therefore, we will further investigate whether it needs to introduce the SS mechanism in the dialogue model in the future study. \par
\begin{table*}[htb]
\centering
\captionsetup{font={normal}}
\caption{Results of ablation experiments.}
\resizebox{\textwidth}{!}{
\renewcommand{\arraystretch}{1.5}
\begin{tabular}{lccccccc}
\hline
\multirow{2}{*}{Model} & \multicolumn{3}{c}{Persona-Chat} & & \multicolumn{3}{c}{DailyDialog} \\
\cline{2-4} \cline{6-8}
 & {F1} & {BLEU-1/2} & {DISTINCT-1/2} &  & {F1} & {BLEU-1/2} & {DISTINCT-1/2} \\
\hline
{Seq2Seq}    & {16.551} & {0.156/0.052} & {0.007/0.017} & & {16.589} & {0.108/0.041} & {0.027/0.117} \\
{Seq2Seq+SS} & {\textbf{16.653}} & {0.143/0.056} & {0.007/0.021}&  & {16.338} & {0.096/0.039} & {0.026/0.093}  \\
{Seq2Seq+RA} & {16.527} & {0.156/\textbf{0.062}} & {\textbf{0.009/0.032}} & & \textbf{{17.510}} & \textbf{{0.154/0.057}} & {0.029/\textbf{0.171}} \\
{Seq2Seq+SS+RA)} & {16.554} & {\textbf{0.163}/0.057} & {0.005/0.010} & & {16.689} & {0.121/0.047} & {\textbf{0.031}/0.144} \\
\hline
{GPT-2}    & {18.790} & {0.177/0.083} & {0.022/0.092} & & {19.095} & {0.146/0.067} & {0.042/0.195} \\
{GPT-2+SS} & {18.264} & {0.168/0.078} & {0.022/0.090}&  & {17.911} & {0.122/0.053} & {0.032/0.135}  \\
{GPT-2+RA} & {19.233} & {0.177/0.083} & {\textbf{0.023/0.103}} & & \textbf{{25.860}} & {0.235/\textbf{0.156}} & \textbf{0.052/0.304} \\
{GPT-2+SS+RA(RAD)} & \textbf{{19.385}} & {\textbf{0.190/0.089}} & {0.022/0.093} & & {24.656} & {\textbf{0.244}/0.140} & {0.043/0.249} \\

\hline
\end{tabular}}%
\label{tab:tab5}%
\end{table*}%

\section{Conclusion}
This paper proposes a new method utilizing ground-truth responses to enhance the performance of open-domain dialogue models. This method provides information to the pre-trained model by learning the implicit pattern between contexts and responses through a response-aware mechanism. Moreover, an improved scheduled sampling technique is used to eliminate exposure bias. Finally, the automatic and human evaluation results validate that our work can improve the quality of generated responses.\par
%

\paragraph{\textbf{Acknoledgements.}}
This work was supported in part by the Fundamental Research Funds for the Central Universities (No. ZYGX2016J096) and the Science and Technology Programs of Sichuan Province of China (No. 23GJHZ0016).\par

%
%
%
\bibliographystyle{splncs04}
\bibliography{mybib1}
%




\end{document}